# Evaluating the overall sensitivity of saliency-based explanation methods.

**Harshinee Sriram**[1] and **Cristina Conati**[1]
The University of British Columbia
{hsriram, conati}@cs.ubc.ca

## Abstract

We address the need to generate faithful explanations of "black box" Deep Learning models. Several tests have been proposed to determine aspects of faithfulness of explanation methods, but they lack cross-domain applicability and a rigorous methodology. Hence, we select an existing test that is model agnostic and is well-suited for comparing one aspect of faithfulness (i.e., sensitivity) of multiple explanation methods, and extend it by specifying formal thresholds and building criteria to determine the overall sensitivity of the explanation method. We present examples of how multiple explanation methods for Convolutional Neural Networks can be compared using this extended methodology. Finally, we discuss the relationship between sensitivity and faithfulness and consider how the test can be adapted to assess different explanation methods in other domains.

## 1 Introduction

Deep Learning (DL) has revolutionized applications in areas including healthcare, transportation, and finance. However, as DL models are inherently "black box" in nature, it is hard to understand their inner decision-making logic, which makes it difficult to trust and validate them (Rudin, 2019). Research in Explainable AI (XAI) addresses this problem by investigating ways to describe how these black box models work with the help of explanations (van der Waa et al., 2021). An explanation acts as an "interface" between humans and a DL model, and it should be at the same time both an accurate proxy of the algorithm and comprehensible to humans (Guidotti et al., 2018). Explanations can be global or local. Global explanation methods provide explanations for how the model makes decisions with an overall view of all the model's components (e.g., model parameters) and their interactions. Thus, these explanation methods attempt to justify the entire decision process of the black-box model at once. Some examples of global explanation methods to explain tabular data include Global Attribution Mapping (GAM) (Ibrahim et al., 2019), Global Aggregations of Local Explanations (GALE) (van der Linden et al., 2019), and Class Model Visualizations (Simonyan et al., 2013). Local explanation methods, on the other hand, provide explanations for what features were considered important for a specific prediction. Examples of local explanation methods to explain image data are Integrated Gradients (Sundararajan et al., 2017), Local Interpretable Model-Agnostic Explanations (Ribeiro et al., 2016), and Attention (de Santana Correia, 2021). In both cases (i.e., global, and local explanations), we want the generated explanation to be faithful i.e., to truly depict the internal decision process that the model underwent for its predictions (Jain & Wallace, 2019; Wiegreffe & Pinter, 2019; Serrano et al., 2019; Jacovi et al., 2020).

In this paper, we focus on the faithfulness of local explanation methods for DL models and, hence, this is what we refer to with the term "explanations". Additionally, any method that uses an algorithm to generate a particular explanation for an input prediction is called an explanation method. There have been attempts to lay down common assumptions for what criteria should be satisfied for an explanation method to be faithful (Jacovi et al., 2020). However, there is still a lack of a unified understanding of what properties are necessary, sufficient, or nice to have when it comes to determining if an explanation method is faithful or not. Several tests have been proposed to determine the faithfulness of explanation methods (Adebayo et al., 2018; Jain et al., 2019; Wiegreffe et al., 2019; Serrano et al., 2019), however, 2 main issues are observed across these works. First, most of these tests have been applied only to specific domains and explanation methods, with minimal information on how they would extend to other domains and explanation methods, making it challenging to analyze and compare results if one must use a different test for every case scenario. Second, these existing works lack a rigorous methodology because they do not specify proper thresholds to determine when an explanation method passes their tests.

In this work, we contribute to the existing work on faithfulness by focusing on one of the tests proposed in the literature and extending it with a comprehensive methodology to ascertain when an explanation method passes or fails this test. This test is the independent parameter randomization (IPR) test proposed by Adebayo et al. (2018) to assess the sensitivity of saliency-based explanation methods for Convolutional Neural Networks (CNNs) for image classification tasks. The saliency-based explanation methods considered in this work provide explanations in the form of saliency maps, namely, images that highlight the pixel-level regions of the input image that contributed most to a prediction. Figure 1 shows the saliency maps generated by the different explanation methods considered by Adebayo et al., (2018) for the classification of the sample input image on the left. The IPR test checks the sensitivity of a given explanation method to perturbations in a CNN as follows: at every turn, it randomizes one parameter layer of the CNN (the other layers retain their original parameters) and generates explanations for predictions from these randomized-layer models. Using Structural Similarity (SSIM) and rank correlation metrics, if the explanations provided for the original prediction for the image from these randomized layer models differ from those by the original model, the explanation method is sensitive to changes made to the model's parameter layers. Sensitivity to such changes is seen as a necessary condition for the faithfulness of an explanation method, because if the explanation method is insensitive to a layer randomization, it is not leveraging that part of the model architecture and, therefore, it cannot be a faithful depiction of the model's decision-making process.

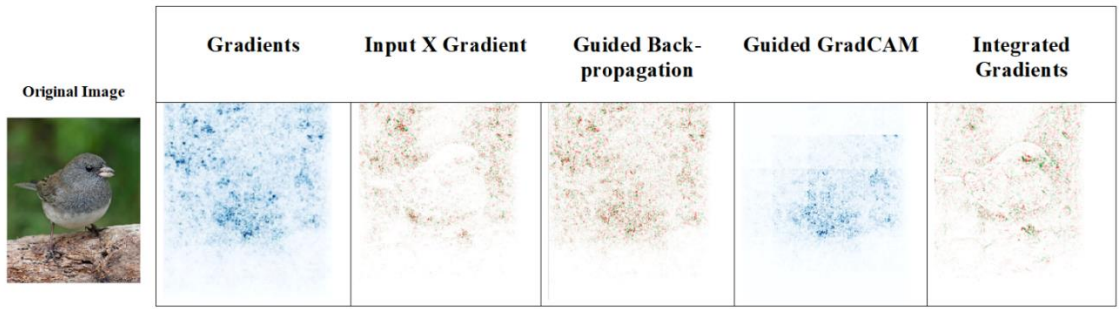

**Figure 1.** Saliency maps generated by the explanation methods listed in the top row (used in the original IPR test by Adebayo et al., 2018), for an example input image fed to the InceptionV3 model architecture.

However, two areas in the IPR test require improvements. First, the comparisons made with similarity metrics are done qualitatively, namely, without a specific threshold that defines when two saliency maps, one from the original model and the other from the randomized layer model are "similar enough". Second, this work does not define criteria that go beyond individual images, namely, criteria to evaluate the sensitivity of a given explanation method over a whole dataset. This makes it challenging to do an overall comparison among multiple explanation methods for a particular model architecture and dataset. However, there are two important merits of the IPR test. First, it is well-suited to compare the sensitivity of multiple saliency-based explanation methods at once for individual images. Second, it is agnostic to the model architecture, so it can be applied to other model architectures and domains as well.

Therefore, in this paper, we extend this notion of sensitivity of an explanation method by addressing the aforementioned limitations of the IPR test in two ways. First, we specify a formal threshold that defines when two saliency maps are "similar enough" to determine when an explanation method is sensitive to the randomization of a given layer. Second, we build on this threshold to define criteria to determine the overall sensitivity of an explanation method with respect to a model architecture for an image and a dataset. We present examples of how the sensitivities of multiple explanation methods for CNNs can be compared using this extended IPR test methodology. Next, we briefly discuss the relationship between the sensitivity of an explanation method and its faithfulness. Finally, we consider the extended utility of our test by providing intuition for how it can be adapted to assess different explanation methods in other domains (for example, natural language processing).

## 2 Related Work

In this paper, we focus on CNNs for an image classification task. The core component of a CNN is a convolutional block, which contains a convolutional layer followed by non-linear activation and a pooling layer. A commonly used activation function is Rectified Linear Unit (ReLU). The convolutional blocks in a CNN form its decision process as these blocks perform the necessary input transformations required to generate the output. Adebayo et al. (2018) performed the IPR test by randomizing each convolutional block. In doing so, they assessed the sensitivity of gradient-based explanation methods to these layer randomizations. Figure 2a shows how saliency maps are generated from gradient-based explanation methods with the partial derivative of the output ($f(x)=y$) with respect to the input ($x$) to understand the relevance of each input feature. Examples of gradient-based explanation methods include Gradients (Simonyan et al., 2013), Input X Gradient (Shrikumar et al., 2016), and Guided Backpropagation (Springenberg et al., 2014).

Another group of explanation methods is perturbation-based. Figure 2b shows how saliency maps are generated from a perturbation-based explanation method that uses a surrogate model. The surrogate model generating the saliency maps is trained with perturbed data and the outputs from the original CNN. Examples of perturbation-based explanation methods are LIME (Ribeiro et al., 2016), DeConvolution Nets (Zeiler et al., 2014), and RISE (Petsiuk et al., 2018).

Collectively, gradient-based, and perturbation-based explanation methods fall under the category of post-hoc explanation methods (Ancona et al., 2017). Post-hoc explanation

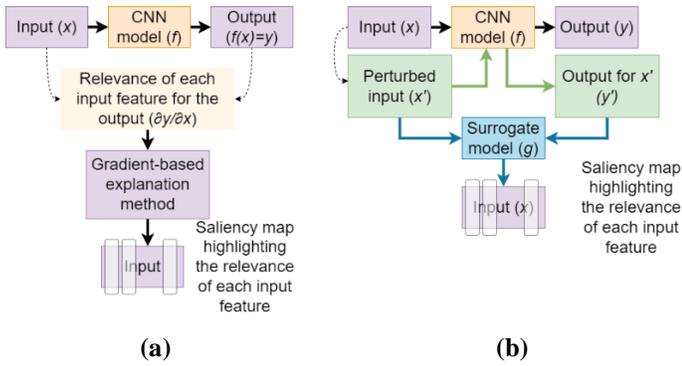

**Figure 2.** An overview of how saliency maps are generated by:
(**a**) gradient-based explanation methods, and
(**b**) perturbation-based explanation methods with a surrogate model methods are externally applied to a fully trained DL model, and they do not interfere with its architecture. In contrast, in-model explanation methods alter the model architecture and affect its performance. An example of an in-model explanation method is attention (Danilevsky et al., 2020).

Most of the work on explanation methods' faithfulness has been performed on attention in the Natural Language Processing (NLP) domain. For example, Jain & Wallace (2019) provided tests to assess to what extent the attention weights in a model measure feature importance and if adversarial attention distributions that yield the same explanations exist. They use BiLSTM models for the tasks of binary text classification, question answering, and natural language inference over multiple datasets such as Stanford Sentiment Treebank, IMDB, 20 Newsgroups, and AG News. Wiegreffe & Pinter (2019) followed up on this work by Jain & Wallace (2019) by using a similar model and a subset of the datasets used by Jain & Wallace (2019) to propose tests to check if adding an attention layer to a model improves its accuracy and by improving the methodology to check if alternate attention distributions that yield similar predictions exist. Serrano et al. (2019) presented tests to determine whether attention weights are good descriptors of the encoded input importance to the model output by determining if erasing the highest attention weights cause a change in a prediction and if there exists a subset of attention weights that significantly influence the model's predictions.

One common limitation of these works is that they do not provide proper thresholds for their methodologies. For example, in the work by Wiegreffe & Pinter (2019), there is no threshold to determine how different should two attention distributions be for them to be considered different or vice versa. Another limitation is that these works do not define a rigorous methodology to perform overall comparisons among multiple explanation methods and datasets. A recent work (Halliwell et al., 2022) reiterates some of these observed issues as the authors state that the lack of a comprehensive evaluation method makes it difficult to evaluate the overall performance of an explanation method.

## 3 The original IPR test methodology

Adebayo et al. (2018) introduced the IPR test as a sanity check to evaluate if an explanation method can even begin to be examined for faithfulness. To generate the randomized layer models from the original model architecture, at each randomization instance, one parameter layer of the original model is independently randomized while all other layers retained their original parameters[1]. For a model with $N$ layers, this generates $N$ randomized layer models after $N$ randomization instances. After each randomization instance, the similarity between the saliency maps produced by the original model and the randomized layer model for a particular input image is determined. This similarity is determined in two ways: first via a visual inspection of the generated saliency maps and then, quantitatively, with three similarity metrics: SSIM (to check for similarities in luminance, contrast, and structure), Spearman rank correlation (to check if the two maps are linearly similar), and Pearson correlation (to check if the two maps are monotonically similar). Once this test is run for all randomized layer models, three different plots for the example input image are generated. Each plot shows the similarity scores between the saliency maps from the original model and the randomized layer models using one of the three similarity metrics. If the saliency maps vary visually and quantitatively with the independent randomization of the layers, it implies that the relevant explanation method is sensitive to these layer randomizations and, hence, can potentially be considered faithful.

Adebayo et al. (2018) report the results of the IPR test for six gradient-based explanation methods. The six explanation methods are Gradients (Simonyan et al., 2013), SmoothGrad (Smilkov et al., 2017), Input X Gradient (Shrikumar et al., 2016), Guided Backpropagation (Springenberg et al., 2014), Guided GradCAM (Selvaraju et al., 2017), and Integrated Gradients (Sundararajan et al., 2017), which have been described in Appendix A. The conclusion from the IPR test is that Gradients and GradCAM are sensitive to parameter layer randomizations. This is because the saliency maps for the example input visibly change after every independent layer randomization, and the similarity between the saliency maps from the randomized layer model architecture and the original model also decrease. On the other hand, Guided GradCAM (Guided GC) and Guided Backpropagation (Guided BP) fail the test because upon visual inspection of the saliency maps from the example input image, it is observed that they are not sensitive to randomizations made to the bottom layers of the model architecture. Lastly, the authors note that

---

[1] In addition to IPR, the authors also propose the Cascading Parameter Randomization (CPR) test. At each parameter layer randomization stage of CPR, the previous layers retain their randomized parameters. CPR informs us of the sensitivity of an explanation method with respect to randomizations made to a block of layers.

We only consider IPR because it is more fine-grained as it tells us to which specific parameter layers is the explanation method sensitive to the randomizations of. We e-mailed Dr. Adebayo to confirm the same.

the saliency maps for Input X Gradient and Integrated Gradients (IG) still visually retain the structure of the input image but quantitatively these saliency maps are dissimilar to those generated by the original model. They do not mention if these two explanation methods pass or fail the test. Additionally, as these evaluations are made with respect to a select few input images, it is unclear if these results would extend to other images in the dataset as well.

## 4 Extending the IPR test methodology.

We extend the IPR test methodology by Adebayo et al. (2018) in three ways:
- By defining an SSIM threshold to determine when two saliency maps are similar
- By determining when an explanation method is sensitive to layer randomizations for an image
- By determining when an explanation method is sensitive to layer randomizations for a dataset

**Notation:** Let:
- $M$ be a model architecture with a set of trained parameter layers $L$
- $L_c$ be the set of layers in M that are critical for its predictions ($L_c \subseteq L$)
- $l_j \in L$ be the trained parameter layers of M
- $l_j'$ be the randomized layer corresponding to a trained parameter layer $l_j$
- $M_L$ be the model with the original layers
- $M_{l_j'}$ be the model with the randomized layer $l_j'$
- $I$ be the set of images i.e., $i_1, i_2, \ldots, i_n \in I$
- $E$ be an explanation method
- $E^i_{M_{l_j'}}$ be the explanation generated by $E$ for the prediction of an image $i$ from model $M_{l_j'}$
- $SSIM(E^i_{M_L}, E^i_{M_{l_j'}})$ be the SSIM between the explanations generated by the explanation method $E$ from the models $M_L$ and $M_{l_j'}$ for the prediction on image $i$

**Defining an SSIM threshold to determine when two saliency maps are similar:** To define a threshold for SSIM, we rely on the mapping between SSIM and human-perceived image similarity identified by Wang et al. (2004). Wang et a (2004) devised this mapping by having human subjects compare low- versus high-resolution images with the help of a 5-point Likert scale rating, where 1 implies low visual similarity and 5 means that they are visually identical. These raw scores are then normalized, rescaled, and mapped to a partitioning of the SSIM (table 1). This mapping has been extensively used in the context of image quality assessment (e.g., Zinner et al., 2010; Zanforlin et al., 2014).

Based on Table 1, we choose 0.99 as the threshold SSIM score to define when two saliency maps are "similar enough". It must be noted that while we do provide a threshold that accounts for comparisons performed with the SSIM metric, our rationale relies on the notion that, to be considered sensitive, the randomized explanation should exhibit a minimal amount of change. It is possible that certain use cases might require a higher threshold because observing a minimal amount of sensitivity would not suffice. In such cases, our test methodology can be easily configured to include this new threshold.

| SSIM | Perceived Image Similarity |
|---|---|
| ≥ 0.99 | Excellent |
| [0.95, 0.99) | Good |
| [0.88, 0.95) | Fair |
| [0.5, 0.88) | Poor |
| < 0.5 | Bad |

**Table 1.** Relationship between SSIM and perceived image similarity (Wang et al., 2004)

*Definition 1* – **Sensitivity of an explanation method $E$ to the randomization of a layer $l_j$ for an image $i$:** For a given E, if the SSIM score between the saliency map generated by a randomized layer model ($E^i_{M_{l_j'}}$) and the saliency map generated by the original model ($E^i_{M_L}$) for the image $i$ is such that $SSIM(E^i_{M_L}, E^i_{M_{l_j'}}) < 0.99$, then E is sensitive to the randomization of the layer $l_j$.

**Determining when an explanation method is sensitive to layer randomizations for an image**: As we now have an SSIM threshold to determine if an explanation method is sensitive to the randomization of a layer or not, we can build on this to define a criterion to establish the sensitivity of an explanation method for a specific image.

*Definition 2* – **Sensitivity of an explanation method E for an image $i$:** We state that $E$ is sensitive to $i$ if the saliency maps generated by $E$ for $i$ with model $M$ are sensitive to the randomizations of all layers ($l_j \in L$) in $M$. That is:

$$E \text{ sensitive to } i \leftrightarrow SSIM(E^i_{M_L}, E^i_{M_{l_j'}}) \leq 0.99 \ \forall \ l_j \in L \quad (1)$$

It should be noted our definition assumes that all layers in $M$ are critical for its predictions (i.e., $L_c = L$). We use this definition for the model architectures that we apply the test on. However, there is some preliminary work that suggests that some layers in the model might be more critical than others (Zhang et al., 2019), although further investigation is required to substantiate the same. In the case that a subset of the layers in the model is known to be critical for its predictions, our definition can be easily adjusted to check if the explanation method is sensitive to these critical layers only. Note that for a given layer $l_j$ and an image $i$, if the $SSIM(E^i_{M_L}, E^i_{M_{l_j'}})$ value is high, then the explanation method $E$ has a low sensitivity to the randomization of $l_j$. Hence, we define the sensitivity $S^i_{l_j}$ of $E$ to the randomization of $l_j$ in model $M$ for $i$ as:

$$S^i_{l_j} = 1 - SSIM(E^i_{M_L}, E^i_{M_{l_j'}}) \quad (2)$$

Using the SSIM threshold from Definition 1, we derive the threshold for sensitivity $S^i_{l_j}$ as:

$$Threshold\left(S^i_{l_j}\right) = 1 - 0.99 = 0.01 \quad (3)$$

We determine the overall sensitivity of $E$ with respect to explanations for $i$ generated by all layer randomizations by averaging the $S^i_{l_j}$ scores of all layers $l \in L$ over the total number of layers $|L|$, that is:

$$S^i = \frac{\sum_{l_j \in L} S^i_{l_j}}{|L|} \quad (4)$$

To determine the threshold for $S^i$, we substitute the value of $S^i_{l_j}$ in (4) with (3), thus:

$$Threshold(S^i) = \frac{\sum_{l_j \in L}(0.01)}{|L|} = \frac{|L| \cdot 0.01}{|L|} = 0.01 \quad (5)$$

**Determining when an explanation method is sensitive to layer randomizations for a dataset:** From (1) we have a criterion for when $E$ is considered sensitive to $i$, which we use to calculate $S^i$ in (4). We extend this reasoning to determine when $E$ is sensitive to a set of images $I$. Thus, in this case, we determine the sensitivity of E to layer randomizations for $I$ by averaging the $S^i$ scores of all images $i \in I$ over the total number of images $|I|$, that is:

$$S^I = \frac{\sum_{i \in I} S^i}{|I|} \quad (6)$$

To determine the threshold for $S^I$, we substitute the value of $S^i$ in (6) with (5), thus:

$$Threshold(S^I) = \frac{\sum_{i \in I}(0.01)}{|I|} = \frac{|I| \cdot (0.01)}{|I|} = 0.01 \quad (7)$$

*Definition 3 – Sensitivity of an explanation method E for a dataset $I$:* We state that an explanation method $E$ is sensitive to a set of images $I$ if: $S^I \geq 0.01$. As SSIM$\in[0,1]$, from (2) and (6), we observe that, for any explanation method $E$, $S^I \in [0,1]$. An $S^I$ score of 0 means that the explanation method is not sensitive to model M for dataset I and a score of 1 means that it has maximum sensitivity.

## 5 Determining the $S^I$ of explanation methods with the extended IPR test.

In this section, we show how our proposed extension to the IPR test can be used to compare and evaluate the sensitivity of different post-hoc explanations methods for DL models for the task of image classification. We use the InceptionV3 model architecture and the ImageNet dataset as this was the setup used by Adebayo et al. (2019) for the original IPR test. In addition, we include the model architectures of VGG16, MobileNetV2, ResNet18, and ResNet50. When it comes to the list of explanation methods, we evaluate Gradients, InputXGradient, Guided Backpropagation, Guided GradCAM, and Integrated Gradients because these were evaluated by the original IPR test as well. To this list, we add DeepLIFT and LIME.

**Dataset:** We use the ImageNet dataset to evaluate the sensitivity of explanation methods with the extended IPR test. The ImageNet dataset was used by Adebayo et al. (2018) for the original IPR test and is a standard dataset used for image classification tasks.

**Model architectures:** We investigate the sensitivity of post-hoc explanation methods in the domain of image classification. Hence, we begin by selecting VGG16 (Simonyan et al., 2014), a CNN that has been investigated as a baseline for numerous image classification tasks. VGG16 comprises of sequential convolutional blocks and a linear weight propagation path. However, VGG16 has many convolutional blocks, which makes it computationally expensive and leads to the *vanishing gradients* problem. There are two solutions to this problem: (1) creating shallower CNNs, and (2) developing a non-linear weight propagation path. InceptionV3 (Szegedy et al., 2016) and MobileNetV2 (Sandler et al., 2018) use the first solution as they break down large convolutional layers into smaller and shallow convolutional layers. The ResNet models (He et al., 2016) use the second solution as their architectures contain residual blocks that propagate the weights in a non-linear path. In these blocks, the weights sent to the non-linear activation layer (i.e., activation ReLU layer) are not determined linearly but, instead, are a combination (addition) of two separate paths. Hence, the gradients are not lost as they travel through multiple paths to reach the next residual block.

**Explanation methods:** With the extended IPR test, we evaluate the sensitivity of both types of post-hoc explanation methods: gradient-based and perturbation-based. The selected gradient-based explanation methods are Gradients (Simonyan et al., 2013), Input X Gradient (Shrikumar et al., 2016), Guided Backpropagation (Springenberg et al., 2014), Guided GradCAM (Selvaraju et al., 2017), DeepLIFT (Shrikumar et al., 2017), and Integrated Gradients (Sundararajan et al., 2017). Gradients, InputXGradient, Guided Backpropagation, Guided GradCAM, and Integrated Gradients were a part of the original IPR test. DeepLIFT is added to this list because recent work provides evidence that it is better at reflecting changes made to the back propagated vector than other gradient-based explanation methods (Sixt et al., 2020). The selected perturbation-based explanation method is LIME (Ribeiro et al., 2016). These explanation methods have been further described in Appendix A.

## 6 Experimental setup and procedure

First, for each model, randomized layer models were generated and stored. Each randomized layer model contained one layer with randomized parameters and the remaining layers retained their original parameters. This led to the creation of 17 randomized layer models for InceptionV3, 13 randomized layer models for VGG16, 19 randomized layer models for MobileNetV2, 8 randomized layer models for ResNet18, and 16 randomized layer models for ResNet50. As the ImageNet dataset has around 14 million images, it is unfeasible to run the test for all images in the dataset for each of the five model architectures. Instead, we perform the test on a set of 900 images from the validation set with a bootstrapped confidence interval of 90%. This means that the set of images is selected by a resampling process that helps with making inferences about an entire population from this sample data with a 90% level of confidence. For each model, at every IPR instance, we generate the sensitivity for all images in the sample dataset. This means that at every instance, for each image, we calculate the SSIM between the saliency map generated by the original model architecture and that generated by the specific randomized layer model architecture. Once the SSIM scores for the explanations of the image have been generated for all layers in the model architecture, we determine if the explanation method is sensitive to this image or not based on the criteria presented in (1). From this result, (3) and (4) are used to calculate the sensitivity of the explanation method for that image. This process is repeated for all images in the sample. After this, with (5), the $S^I$ value of the explanation method is calculated. Once this process is completed for all model architectures and explanation methods, we compare these $S^I$ values to draw inferences. To generate the post-hoc explanations, we use the Captum library developed by Facebook AI for PyTorch (Kokhlikyan et al., 2020). The test was performed on a system with NVIDIA GeForce RTX 2060 and performing the test for each model architecture took about 1 week to complete.

## 7 Results

Figure 3 shows the $S^I$ scores of each explanation method across all model architectures. The horizontal dashed red line indicates the $S^I$ threshold of 0.01, derived from (8). We observe that all explanation methods pass this threshold and, hence, all of them pass the extended IPR test. This contrasts some of the results reported by Adebayo et al. (2018) where, for the InceptionV3 model architecture on the ImageNet dataset, Guided GC and Guided BP fail their test (the original IPR test) because they are not sensitive to randomizations made to the bottom layers of the model architecture. However, they do not provide a reason for this observed phenomenon. In addition, Dr. Adebayo redirected us to the work by Sixt et al. (2020) which mentions that the implementations of the Guided GC and Guided BP algorithms for the original IPR test were flawed, resulting in the insensitive saliency maps. Hence, it is unclear if Adebayo et al. (2018) would have drawn the same conclusions for Guided BP and Guided GC if their implementation of these algorithms were correct and resulted in saliency maps that change with every randomization instance. We perform a correlation analysis with Spearman and Pearson correlation for all explanation methods to determine if certain groups of explanation methods have significantly high correlations. We decided to use both correlation metrics because Spearman correlation considers the ranks of the $S^I$ scores to tell us if there is a monotonic relationship between them whereas Pearson correlation considers the true $S^I$ scores to tell us if there is a linear relationship between them. Appendix B shows the complete results of our correlation analysis. We notice a high correlation for the two groups of explanation methods. The first group includes Guided BP and Guided GC whereas the second group includes DeepLIFT, InputXGradient, and IG. Hence, we first discuss the possible reasons behind the high correlation in $S^I$ scores between Guided BP and Guided GC. Next, we discuss the reasons behind the observed high correlation in $S^I$ scores for DeepLIFT, InputXGradient, and IG. Finally, we discuss the rest of the explanation methods i.e., Gradients and LIME.

**Results for Guided BP and Guided GC:** In addition to the high correlation in $S^I$ scores for Guided BP and Guided GC, we also notice a high variance in these scores across all five model architectures. In contrast, other explanation methods have consistently high $S^I$ scores. Guided GC builds up on Guided BP by generating saliency maps through an element-wise product of the Guided BP saliency weights and the saliency weights from the last convolutional layer of the model, which provides high-resolution and class-discriminative explanations. Hence, this similarity between the $S^I$ trends of Guided BP and Guided GC is unsurprising and Guided BP is the common factor for this observed $S^I$ variance. We notice that these two explanation methods have the lowest $S^I$ scores for VGG16, ResNet18, and ResNet50. For the VGG16, this could possibly be due to the high number of ReLU components in its architecture. VGG16 has the highest number of ReLU components among all the model architectures because it contains ReLU components in both convolutional blocks and fully connected blocks. Guided BP ignores any effects of ReLU layers; hence, it is not affected by the randomization of these layers, leading to a low $S^I$ score. The same reasoning applies to Guided GC. For ResNet models, the low $S^I$ scores could be due to the residual blocks in their architectures resulting in a non-linear weight-propagation path. As a result, Guided BP fails to directly obtain the saliency weights for its explanations, resulting in low $S^I$. In contrast, Guided BP has higher sensitivity for InceptionV3 and MobileNetV2. This could be attributed to similarities in convolution operations performed by the parameter layers in both architectures. As mentioned in Section 5, InceptionV3 and MobileNetV2 factorize larger convolutional layers into smaller convolutional segments. Hence, the resulting shallower networks are possibly more sensitive to these layer randomizations.

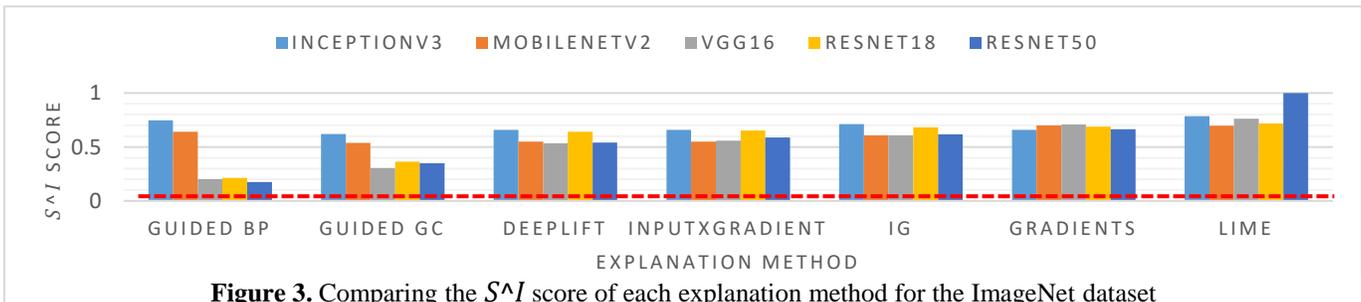

**Figure 3.** Comparing the $S^I$ score of each explanation method for the ImageNet dataset

**Results for DeepLIFT, InputXGradient, and IG:** We observe a significantly high correlation in the $S^I$ scores for DeepLIFT, InputXGradient, and IG. This observed similarity in their behavior is consistent with an analysis presented by Ancona et al. (2017), who provide a mathematical justification for the same. The authors show that DeepLIFT and InputXGradient work similarly for a model architecture with only non-linearities (such as ReLU elements) and no additive biases, which is the case for all five model architectures considered here. InputXGradient and IG are related because they both compute the partial derivatives of the output with respect to each input feature. Finally, DeepLIFT and IG are related as well because while IG computes the average partial derivative of each feature as the input varies from a baseline to its final value, DeepLIFT approximates this quantity in a single step by replacing the gradient at each nonlinearity with its average gradient.

**Results for Gradients:** The explanation method of Gradients exhibits high sensitivity and small variance like the three explanation methods that we just discussed, but its $S^I$ scores are not significantly correlated to them. Regardless, due to its high consistency in $S^I$ scores, one cannot go wrong if they select Gradients as an explanation method for these model architectures if high sensitivity is valued.

**Results for LIME:** Our results show that LIME has consistently high $S^I$ scores, comparable to Gradients, except for ResNet50, for which it has an $S^I$ score of 1.0 (the highest possible $S^I$ score). We hypothesize that this high sensitivity can be attributed to the surrogate model that generates the explanations. The surrogate model appears to be highly sensitive to layer randomizations made to the original CNN-based model architecture. This could be because the surrogate model is a local and simpler approximation of the original CNN-based model architecture. However, further investigation is required to understand the reason behind this observation.

From these results, we learn how certain aspects of an explanation method, or a model architecture can affect its $S^I$ score. This extended test provides a comprehensive overview of the overall sensitivity of a list of explanation methods. Based on these results, selecting an ideal explanation method would depend on what level of sensitivity is required for the task at hand.

# 8 The relationship between sensitivity towards parameter layers and faithfulness

Our test, as discussed in the beginning, should be seen as a necessary condition for an explanation method to be considered faithful. As such, even when an explanation method passes this test, other aspects need to be tested and evaluated to understand if it fulfills other aspects of faithfulness (Jacovi et al., 2020). This is what other tests in the literature have been looking at. These criteria include (i) a faithful explanation method must provide similar explanations for similar inputs, and that (ii) a faithful explanation method must be robust to small perturbations made to the parameter layers and the inputs. The tests by Jain & Wallace (2019) assess the potential of an explanation method to be faithful using the first criterion, whereas the work by Alvarez-Melis et al. (2018) deals with the second criterion by assessing the robustness of explanation methods with the Lipschitz Continuity Metric.

# 9 Conclusion and future work

In this work, we extend the IPR test in two ways. First, we define a threshold to determine when two saliency maps are similar. Second, we provide a methodology to summarize the overall sensitivity of an explanation method for a model architecture and dataset. We show how the extended test can be used to compare explanation methods based on their $S^I$ scores. The extended IPR test does not measure the magnitude of perturbations because it is meant to inform if the saliency maps change if at all, to layer randomizations. Future work would take the magnitude of perturbations into account. It would be interesting to observe how different reference inputs affect the sensitivities of IG and DeepLIFT. In this work, we use an image where all pixels are zero as a reference but there are other types of reference inputs too, namely, maximum distance, blurred, uniform, and Gaussian (Sturmfels et al., 2020). This is interesting because there is evidence stating that using a blank image as a reference input for DeepLIFT may lead to biased saliency weights compared to those from a randomly sampled distribution (Lundberg et al., 2017; Chen et al., 2021). For LIME, it would be interesting to see how changing the hyper-parameters (such as changing the type of surrogate model) may affect its sensitivity.

Additionally, we would like to apply the extended test to the in-model explanation method of attention. When used in the visual domain, explanations generated by attention are in the form of saliency maps. Hence, the presented version of the extended IPR test can be directly applied to assess the sensitivity of attention in the visual domain. We would also like to assess the sensitivity of attention in other domains such as NLP. In these domains, the attention weights are probability distributions. Hence, to determine the similarity between probability distributions, we can use Jensen-Shannon Divergence (Fuglede et al., 2004) instead of SSIM. Setting a threshold for Jensen Shannon Divergence, in this case, can be determined by further research.

# Appendix

## A. Explanation method descriptions

### A.1 Gradient-based explanation method(s)
- **Gradients:** Gradients generate a saliency map with the vanilla gradients obtained directly from the partial derivative of the output with respect to the input (Simonyan et. al., 2013).
- **InputXGradient:** Here, the partial derivative is multiplied by the input feature values (Shrikumar et. al., 2016).
- **Guided Backpropagation (Guided BP):** Here, the backpropagation of ReLU functions is overridden when computing the partial derivative (Springenberg et. al., 2014). This is done so that only non-negative gradients are back-propagated because back-propagating non-negative and negative gradients results in an interference pattern that makes the resulting gradients noisy.
- **Guided GradCAM (Guided GC):** The saliency maps here are generated with the help of an element-wise product of the guided backpropagation saliency weights and the saliency weights derived from the last convolutional layer of the model (Selvaraju et. al., 2017). The saliency weights derived from the last convolutional layer of the model are used to generate explanations for the vanilla GradCAM explanation method (Selvaraju et al., 2017). Hence, Guided GC explanations combine the high-resolution nature of Guided BP along with the class-discriminative advantages that GradCAM provides.
- **DeepLIFT:** The explanations generated by DeepLIFT require a blank image (i.e., an image where all pixels are zero) as reference. This is because this explanation method uses this blank image to determine the input features that the model would highlight by default. With this knowledge, it provides explanations for a required image after removing the default highlighted features (Shrikumar et. al., 2017).
- **Integrated Gradients (IG):** Similar to DeepLIFT, IG (Sundararajan et. al., 2017) generates saliency maps with the help of a blank image. However, in this case, the difference between the features highlighted for the required image and those for the blank image are integrated.

### A.2 The LIME perturbation-based explanation method:
What makes LIME (Ribeiro et. al., 2016) different from the other explanation methods described above is that saliency maps generated by LIME are not derived from the original DL model. Instead, they are generated with the help of a surrogate model, which is a simpler and arguably inherently interpretable model (such as a linear model). To figure out what parts of the input are contributing to the prediction, the input is perturbed around its neighbourhood and the DL model's predictions on these inputs is monitored. Then, these perturbed data points are weighed by their proximity to the original example. This is used for the learning process of the surrogate model. We use the default Lasso model as our surrogate model, which is a linear model trained with an L1 prior as the regularizer.

## B. Correlation analysis of the $S^I$ scores of all 7 explanation methods

Tables 2 and 3 show the Spearman and Pearson correlation values respectively for all 7 explanation methods.

**Table 2.** Spearman correlation among the $S^I$ scores of all explanation methods for all model architectures

|  | Guided BP | Guided GC | DeepLIFT | InputXGradient | IG | Gradients | LIME |
|---|---|---|---|---|---|---|---|
| **Guided BP** | 1 | 0.9 (*) | 0.8 (ns) | 0.3 (ns) | 0.3 (ns) | -0.3 (ns) | -0.4 (ns) |
| **Guided GC** | 0.9 (*) | 1 | 0.9 (ns) | 0.4 (ns) | 0.4 (ns) | -0.6 (ns) | -0.2 (ns) |
| **DeepLIFT** | 0.8 (ns) | 0.9 (ns) | 1 | 0.7 (ns) | 0.7 (ns) | -0.7 (ns) | -0.1 (ns) |
| **InputXGradient** | 0.3 (ns) | 0.4 (ns) | 0.7 (ns) | 1 | 1 (***) | -0.8 (ns) | 0.5 (ns) |
| **IG** | 0.3 (ns) | 0.4 (ns) | 0.7 (ns) | 1 (***) | 1 | -0.8 (ns) | 0.5 (ns) |
| **Gradients** | -0.3 (ns) | -0.6 (ns) | -0.7 (ns) | -0.8 (ns) | -0.8 (ns) | 1 | -0.6 (ns) |
| **LIME** | -0.4 (ns) | -0.2 (ns) | -0.1 (ns) | 0.5 (ns) | 0.5 (ns) | -0.6 (ns) | 1 |
| **(ns = not supported, * = p ≤ 0.05, ** = p ≤ 0.01, *** = p ≤ 0.001)** | | | | | | | |

**Table 3.** Pearson correlation among the $S^I$ scores of all explanation methods for all model architectures

|  | Guided BP | Guided GC | DeepLIFT | InputXGradient | IG | Gradients | LIME |
|---|---|---|---|---|---|---|---|
| **Guided BP** | 1 | 0.9843 (**) | 0.4187 (ns) | 0.1809 (ns) | 0.3962 (ns) | -0.2681 (ns) | -0.3885 (ns) |
| **Guided GC** | 0.9843 (**) | 1 | 0.5138 (ns) | 0.3082 (ns) | 0.4971 (ns) | -0.4129 (ns) | -0.3004 (ns) |
| **DeepLIFT** | 0.4187 (ns) | 0.5138 (ns) | 1 | 0.949 (*) | 0.9847 (**) | -0.4930 (ns) | -0.3175 (ns) |
| **InputXGradient** | 0.1809 (ns) | 0.3082 (ns) | 0.949 (*) | 1 | 0.9668 (**) | -0.6291 (ns) | -0.048 (ns) |
| **IG** | 0.3962 (ns) | 0.4971 (ns) | 0.9847 (**) | 0.9668 (**) | 1 | -0.591 (ns) | -0.1874 (ns) |
| **Gradients** | -0.2681 (ns) | -0.4129 (ns) | -0.4930 (ns) | -0.6291 (ns) | -0.5910 (ns) | 1 | -0.6071 (ns) |
| **LIME** | -0.3885 (ns) | -0.3004 (ns) | -0.3175 (ns) | -0.048 (ns) | -0.1874 (ns) | -0.6071 (ns) | 1 |
| **(ns = not supported, * = p ≤ 0.05, ** = p ≤ 0.01, *** = p ≤ 0.001)** | | | | | | | |